# XNLI 2.0: Improving XNLI dataset and performance on Cross Lingual Understanding (XLU)

Ankit Kumar Upadhyay
*Dept. of Computer Science and Engineering*
*JSS Academy of Technical Education*
Bengaluru, India
ankitkupadhyay@jssateb.ac.in

Harsit Kumar Upadhya
*Dept. of Information Science and Engineering*
*JSS Academy of Technical Education*
Bengaluru, India
1js19is039@jssateb.ac.in

*Abstract*—Natural Language Processing systems are heavily dependent on the availability of annotated data to train practical models. Primarily, models are trained on English datasets. In recent times, significant advances have been made in multilingual understanding due to the steeply increasing necessity of working in different languages. One of the points that stands out is that since there are now so many pre-trained multilingual models, we can utilize them for cross-lingual understanding tasks. Using cross-lingual understanding and Natural Language Inference, it is possible to train models whose applications extend beyond the training language. We can leverage the power of machine translation to skip the tiresome part of translating datasets from one language to another. In this work, we focus on improving the original XNLI dataset by re-translating the MNLI dataset in all of the 14 different languages present in XNLI, including the test and dev sets of XNLI using Google Translate. We also perform experiments by training models in all 15 languages and analyzing their performance on the task of natural language inference. We then expand our boundary to investigate if we could improve performance in low-resource languages such as Swahili and Urdu by training models in languages other than English.

## I. Introduction

Natural Language Processing systems typically rely on annotated data in one language on which it is trained and performs the task in that language only. As the world is globalized and the products developed in one country are being shipped across boundaries, it has become essential that our language system not only understands the language on which it is trained but also other languages of the user that interacts with the product. Often, it is not possible to annotate data in all possible languages.

Through Cross-Lingual Language Understanding (XLU), it is now possible to transfer learning from one natural language to another, especially when we do not have much data available for the target language. This paper focuses on Natural Language Inference (NLI), in which we are given a pair of sentences named premise and hypothesis. The system has to predict if the hypothesis entails the premise, contradicts it or is neutral. We make use of the Cross-lingual Natural Language Inference Corpus (XNLI) that consists of a crowdsourced collection of 5000 test and 2500 development pairs in English, French, Spanish, German, Greek, Bulgarian, Russian, Turkish, Arabic, Vietnamese, Thai, Chinese, Hindi, Swahili, and Urdu. The XNLI corpus consists of a total of 112, 500 annotated pairs.

While going through the original machine translations of MNLI to the Hindi language, we observed a lot of discrepancies. These wrong translations could severely impact the performance of a model. Cloud translation tools of present times are significantly more advanced than a few years before. As a result, it eliminates the need for proficiency in multiple languages to translate the dataset correctly. We have used Google Translate to freshly translate the MNLI dataset to 14 different language training datasets. Our paper specifically talks about the drawbacks and deficiencies of the current XNLI corpus and the impact that better translation could have on the model performance. We will be utilizing the new XNLI dataset, called XNLI 2.0 for reporting accuracy metrics.

XNLI corpus is focused on test and development example pairs, and it is used for the evaluation of cross-lingual sentence understanding, where models are trained on one of the languages and then tested in other languages. Most of the models until now have been trained only in English and evaluated in other languages. However, in this paper, we will try to focus on performing cross-lingual transfer learning by training separate models on each of the 15 languages and then comparing the accuracy to determine which language model performs better during evaluation on the test set. For this experiment, we machine-translate the MNLI dataset to 14 different languages in the XNLI dataset. We treat these machine-translated datasets as separate training sets for the particular language.

Cross-lingual language understanding has been of utmost importance also because of its use in understanding low-resource languages. Languages like Urdu and Swahili are classified as low-resource because there are few available or annotated datasets in these languages, making it difficult to train a model for performing tasks in these languages. In this paper, we also perform multiple experiments to determine which high-resource languages could better help learn these low-resource languages in the XNLI corpus.

| Sr. No. | Translation Language | English Premise/Hypothesis | Original translation | New translation |
|---|---|---|---|---|
| 01 | Hindi | The tennis shoes have a range of prices | टेनिस के जूते की कीमतों की सीमा होती है । | टेनिस जूते की कीमतों की एक श्रृंखला है। |
| 02 | Hindi | One of our number will carry out your instructions minutely. | हमारे एक नंबर में से एक आपके निर् ○ देशों को मिनटों में ले जाएगा . | हमारा एक नंबर आपके निर्देशों को सूक्ष्मता से पूरा करेगा। |
| 03 | Hindi | my walkman broke so i 'm upset now i just have to turn the stereo up real loud | मेरे वॉकमैन टूट गए तो मैं परेशान हूँ अब मुझे बस स ○ टीरियो को असली जोर देना होगा | मेरा वॉकमैन टूट गया इसलिए मैं अब परेशान हूं मुझे बस स्टीरियो को जोर से चालू करना है |
| 04 | Hindi | He turned and smiled at Vrenna | वह दिया पर मुस ○ कराया और मुस ○ कुराया । | वह मुड़ा और वृन्ना की ओर मुस्कुराया। |
| 05 | German | my walkman broke so i 'm upset now i just have to turn the stereo up real loud | Mein Walkman ist kaputt , also bin ich sauer , jetzt muss ich nur noch die Stereoanlage ganz laut drehen . | Mein Walkman ist kaputt gegangen, also bin ich verärgert, jetzt muss ich nur noch die Stereoanlage richtig laut aufdrehen |
| 06 | French | my walkman broke so i 'm upset now i just have to turn the stereo up real loud | Mon Walkman S' est cassé alors je suis en colère maintenant je dois juste tourner la stéréo très fort | mon baladeur est tombé en panne donc je suis contrarié maintenant je n'ai plus qu'à mettre la stéréo à fond |
| 07 | Swahili | my walkman broke so i 'm upset now i just have to turn the stereo up real loud | My broke my sasa sasa sasa up up up up up up up | mtembezaji wangu alivunjika kwa hivyo nimekasirika sasa lazima nipandishe stereo kwa sauti kubwa |

Fig. 1. Sample discrepancies in the original translations of the MNLI dataset.

The paper is organized as follows: We will do a literature review of works relating to the XNLI corpus in cross-lingual learning. In section 3, we try to present the steps undertaken in concluding the mistranslations in the XNLI corpus and building up the newly translated train, test, and dev sets. Section 4 focuses on describing metrics and doing a comparative study, reporting training details, analysis, and results. Finally, in the last section, we present concluding remarks and ways in which it can impact or enhance future works.

## II. RELATED WORK

**XNLI** As we know, most natural language processing systems rely on annotated or labelled data. This data is usually in a single language, so the system can only perform the task in that language. However, in practice, systems need to be able to handle inputs in many languages. It is difficult because annotating data in all languages a system might encounter is nearly impossible. In the paper (Conneau et al., 2018), the authors introduce a new benchmark for evaluating natural language processing systems. The benchmark, Cross-lingual Natural Language Inference corpus, or XNLI, consists of 7500 human-annotated development and test examples in the format of NLI three-way classification: premise, hypothesis, and label. It was then translated by employing professional translators in 14 other languages, French, Spanish, German, Greek, Bulgarian, Russian, Turkish, Arabic, Vietnamese, Thai, Chinese, Hindi, Swahili, and Urdu. These languages comprise several language families and contain two low-resource languages, Swahili and Urdu.

**Unsupervised Multilingual Word Embeddings** Words from many languages are represented by multilingual word embeddings (MWEs) in a single distributional vector space. The acquisition of multilingual embeddings using unsupervised MWE (UMWE) methods does not require cross-lingual supervision, which is a considerable improvement over conventional supervised methods and creates a wealth of new opportunities for low-resource languages. However, prior work for learning UMWEs only uses a number of unsupervised bilingual word embeddings (UBWEs) that have been trained individually to produce multilingual embeddings. The interdependencies that exist between numerous languages are not taken advantage of by these methods. A completely unsupervised approach is suggested for learning MWEs1 that directly takes advantage of the relationships between all language pairs to remedy the aforementioned shortcoming. In experiments on cross-lingual word similarity and multilingual word translation, the model significantly outperforms earlier methods and even bests supervised methods developed using cross-lingual resources (Chen and Cardie et al., 2018).

**GLUE** Technology for natural language understanding (NLU) must be able to handle language in a way that is not restricted to a specific task or dataset if it is to be useful. The General Language Understanding Evaluation (GLUE) benchmark (Wang et al., 2019) is established in order to achieve this goal. GLUE is created to reward and encourage models that share generic language knowledge across tasks by providing tasks with little to no training data. A collection of specially created diagnostic tests is also part of GLUE, allowing for thorough linguistic analysis of models. Evaluation of baselines based on existing transfer learning methodologies reveals that multi-task training on all tasks outperforms training a separate model for each task. The best model's poor absolute performance, on the other hand, highlights the need for more advanced generic NLU systems.

TABLE I
ACCURACY OF MODELS FINE-TUNED ON EACH OF THE 15 LANGUAGES AND TESTED ON ORIGINAL XNLI TEST SET. TRAINING SET IS SAME FOR EACH LANGUAGE MODEL I.E., NEWLY TRANSLATED MNLI DATASET IN RESPECTIVE LANGUAGES.

| Models/Training Language | ar | bg | zh | en | fr | de | el | hi | ru | es | sw | th | tr | ur | vi | Average |
|---|---|---|---|---|---|---|---|---|---|---|---|---|---|---|---|---|
| Conneau 2020 XLM-R Base | 73.8 | 79.6 | 76.7 | 85.8 | 79.7 | 78.7 | 77.5 | 72.4 | 78.1 | 80.7 | 66.5 | 74.6 | 74.2 | 68.3 | 76.5 | 76.21 |
| English | 71.1 | 76.3 | 72.2 | 83.7 | 76.8 | 75.3 | 74.9 | 68.6 | 74.5 | 77.5 | 65 | 70.6 | 71.7 | 65 | 73.4 | 73.11 |
| Hindi | 72.6 | 77.9 | 75.3 | 81.4 | 77.4 | 76.9 | 76.3 | 73.2 | 76.1 | 78 | 66.6 | 73.3 | 73.7 | 69.6 | 75.8 | 74.94 |
| French | 71.9 | 77.2 | 74 | 82.7 | 78.6 | 77.2 | 76.1 | 70.6 | 75.7 | 78.6 | 65.8 | 71.5 | 72.7 | 67 | 75 | 74.31 |
| Spanish | 72.9 | 78.3 | 74.7 | 82.7 | 77.9 | 77 | 76.3 | 70.8 | 76.4 | 79.9 | 65.7 | 72.1 | 72.7 | 68.1 | 75.7 | 74.75 |
| Urdu | 72.5 | 78.2 | 74.7 | 81.3 | 77.7 | 76.7 | 76.4 | 72.4 | 76 | 77.9 | 66.6 | 74 | 74.5 | 69.8 | 75.9 | 74.97 |
| Chinese | 72.6 | 77.3 | 76.7 | 82.3 | 77.5 | 76.8 | 76.1 | 71.6 | 75.4 | 78.4 | 66.5 | 72.9 | 73.7 | 67.8 | 76.4 | 74.8 |
| German | 73 | 78.5 | 74.6 | 83.2 | 78.3 | 78.7 | 76.8 | 71.2 | 76.4 | 79.2 | 66.5 | 73 | 73.8 | 66.9 | 75.6 | 75.05 |
| Greek | 73.8 | 78.6 | 74.5 | 82.5 | 77.9 | 77.7 | 77.8 | 71.7 | 76.2 | 79.3 | 66.6 | 73 | 73.1 | 67.8 | 75.8 | 75.09 |
| Bulgarian | 73.1 | 79.5 | 74.9 | 82.9 | 78.3 | 78.1 | 76.7 | 71.7 | 77.5 | 78.9 | 66.5 | 72.2 | 73.3 | 68 | 75.8 | 75.16 |
| Thai | 72.9 | 78.1 | 75.2 | 82.1 | 78.1 | 77.2 | 76.9 | 71.4 | 76.2 | 78.8 | 67.4 | 76 | 73.8 | 68.8 | 76.5 | 75.29 |
| Russian | 73 | 78.4 | 75.1 | 82.2 | 78 | 77.1 | 77.6 | 71.1 | 76.1 | 78.9 | 67.1 | 72.2 | 73.1 | 67.9 | 76.3 | 74.94 |
| Vietnamese | 72.5 | 77.4 | 75.1 | 81.8 | 77.2 | 76.2 | 75.4 | 70.9 | 75.2 | 78.4 | 66.5 | 72.1 | 72.8 | 67.7 | 77.5 | 74.45 |
| Turkish | 72.4 | 77.5 | 74.8 | 81.9 | 77.3 | 76.3 | 76 | 72 | 75.8 | 78.6 | 66.6 | 72.5 | 75 | 68.4 | 75.6 | 74.71 |
| Arabic | 74.5 | 77.8 | 74.9 | 81.6 | 77.6 | 76.9 | 76.5 | 71.8 | 76.5 | 78.3 | 66.2 | 73.2 | 72.9 | 68.5 | 76.2 | 74.89 |
| Swahili | 72 | 77.1 | 74.2 | 80.1 | 75.9 | 75.7 | 75.3 | 71 | 74.8 | 77.4 | 70.8 | 72.5 | 72.5 | 67.6 | 75.4 | 74.15 |

TABLE II
ACCURACY OF MODELS FINE-TUNED ON EACH OF THE 15 LANGUAGES AND TESTED ON XNLI 2.0 TEST SET. TRAINING SET IS SAME FOR EACH LANGUAGE MODEL I.E., NEWLY TRANSLATED MNLI DATASET IN RESPECTIVE LANGUAGES.

| Models/Training Language | ar | bg | zh | en | fr | de | el | hi | ru | es | sw | th | tr | ur | vi | Average |
|---|---|---|---|---|---|---|---|---|---|---|---|---|---|---|---|---|
| Conneau 2020 XLM-R Base | 73.8 | 79.6 | 76.7 | 85.8 | 79.7 | 78.7 | 77.5 | 72.4 | 78.1 | 80.7 | 66.5 | 74.6 | 74.2 | 68.3 | 76.5 | 76.21 |
| English | 74.4 | 79 | 76.8 | 83.7 | 78.7 | 78.3 | 76.6 | 75.1 | 78.9 | 79.8 | 68.7 | 73.8 | 74.8 | 73.7 | 77.2 | 76.63 |
| Hindi | 75.6 | 79.5 | 77.6 | 81.4 | 79.7 | 78.4 | 78.1 | 79.5 | 79.9 | 79.4 | 69.7 | 76.1 | 75.9 | 76.4 | 79 | 77.74 |
| French | 74.5 | 79.5 | 77.3 | 82.7 | 80.7 | 79.1 | 78.1 | 75.8 | 79.5 | 79.9 | 69.7 | 74.4 | 75.5 | 75.3 | 78.4 | 77.36 |
| Spanish | 75.6 | 80.1 | 78 | 82.7 | 80.4 | 79.3 | 78 | 76.5 | 80.2 | 81.5 | 69.9 | 74.8 | 76 | 75.8 | 78.5 | 77.81 |
| Urdu | 74.8 | 79.9 | 77.8 | 81.3 | 79.5 | 78.1 | 78 | 77.7 | 79.3 | 79.9 | 70.2 | 76.8 | 76.4 | 78.4 | 78.6 | 77.78 |
| Chinese | 75.2 | 79.8 | 80.8 | 82.3 | 79.5 | 78.9 | 77.4 | 77 | 80 | 79.9 | 69.8 | 75.7 | 76.5 | 76.1 | 78.9 | 77.85 |
| German | 76.6 | 80.2 | 77.6 | 83.2 | 80.3 | 81 | 77.8 | 76.7 | 80.3 | 80.9 | 70.4 | 75.1 | 76.7 | 76.5 | 79 | 78.15 |
| Greek | 76.3 | 80.5 | 77.9 | 82.5 | 79.7 | 79.2 | 79.5 | 76.7 | 79.9 | 80.9 | 70.8 | 75.6 | 75.3 | 75.7 | 78.7 | 77.95 |
| Bulgarian | 75.6 | 81.2 | 78.4 | 82.9 | 80.5 | 79.9 | 78.1 | 76.9 | 80.9 | 80.8 | 69.7 | 75.3 | 76.1 | 75.9 | 79.2 | 78.09 |
| Thai | 75.9 | 79.5 | 78.6 | 82.1 | 79.5 | 79.3 | 78.2 | 75.6 | 79.5 | 80.9 | 70.7 | 78.4 | 75.7 | 74.6 | 79.4 | 77.86 |
| Russian | 75.9 | 79.4 | 77.5 | 82.2 | 79.7 | 78.7 | 79.1 | 76.2 | 79.7 | 80.3 | 70.1 | 75.7 | 75.2 | 75.1 | 78.5 | 77.55 |
| Vietnamese | 75.6 | 79.4 | 78.2 | 81.8 | 79.1 | 78.2 | 77.4 | 76.1 | 78.6 | 79.5 | 69.5 | 75.4 | 75.4 | 75.6 | 80.1 | 77.33 |
| Turkish | 75.2 | 78.8 | 77.9 | 81.9 | 79.3 | 77.8 | 78 | 77.4 | 79.6 | 79.7 | 69.5 | 75.3 | 79.1 | 76.7 | 78.2 | 77.66 |
| Arabic | 78.4 | 79.5 | 77.9 | 81.6 | 79.2 | 78.9 | 77.8 | 76.3 | 79.5 | 79.8 | 69.8 | 75.5 | 76.3 | 75.8 | 79.1 | 77.69 |
| Swahili | 75.1 | 78.5 | 76.8 | 80.1 | 78.4 | 77.7 | 77.2 | 74.9 | 78.5 | 78.4 | 75.9 | 74.8 | 75.2 | 75 | 78.3 | 76.99 |

**XLM-R** It is a transformer-based multilingual masked language model trained on 2.5 TB of newly created clean CommonCrawl data in 100 languages (Conneau et al., 2020). It was used to obtain state-of-the-art performance on cross-lingual classification, sequence labelling and question answering. Models that are pretrained on many languages can be effective in cross-lingual understanding tasks. These models are called multilingual masked language models (MLM). For Cross-lingual understanding, the XLM-R model outperforms the XLM-100 and mBERT models by 10.2% and 14.6% average accuracy. On the Swahili and Urdu low-resource languages, XLM-R outperforms XLM-100 by 15.7% and 11.4%, and mBERT by 23.5% and 15.8%. XLM-R outperforms the Unicoder and XLM models by a 5.5% and 5.8% average accuracy, respectively.

**Cross-Lingual Retrieval Augmented Prompt for Low-Resource Languages** Recent empirical cross-lingual transfer experiments have demonstrated the robust multilinguality of Multilingual Pretrained Language Models (MPLMs). In this research (Nie et al., 2022), the Prompts Augmented by Retrieval Crosslingually (PARC) pipeline is presented to enhance the context with semantically related sentences retrieved from a high-resource language (HRL) as prompts, hence improving the zero-shot performance on low-resource languages (LRLs). With multilingual parallel test sets across 10 LRLs covering 6 language families, PARC increases the zero-shot performance on three downstream tasks (binary sentiment classification, topic categorization, and natural language inference) in both unlabeled settings (+5.1%) and labelled settings (+16.3%).

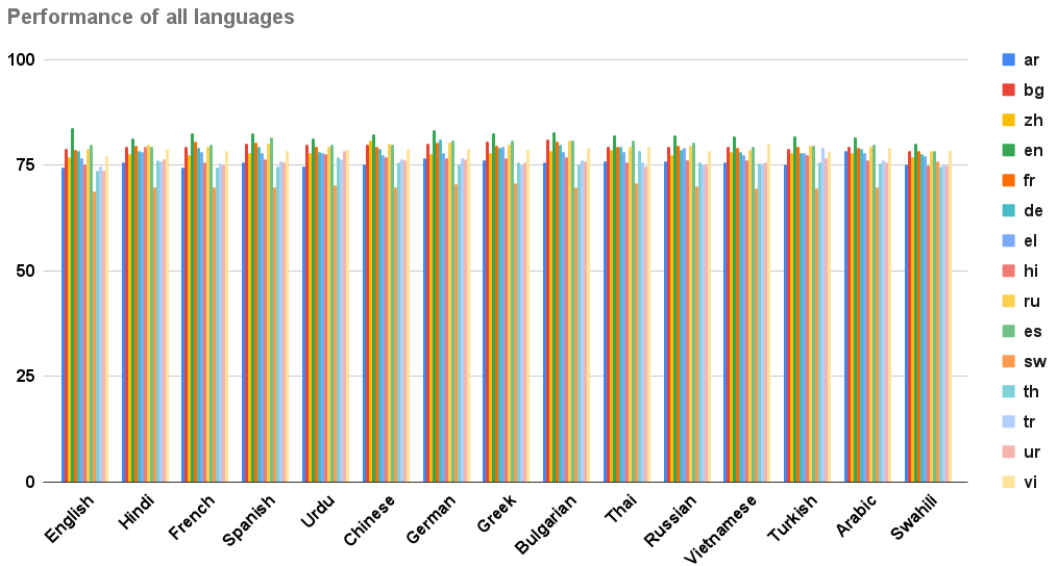

Fig. 2. Horizontal axis denotes the languages on which model is trained and the vertical axis denotes different accuracies for the model tested on all 15 languages in the test set. Different color labels are used to denoted models tested in the respective languages.

Additionally, PARC-labeled surpasses the baseline for fine-tuning by 3.7%. The similarity between the high- and low-resource languages and cross-lingual transfer performance are revealed to be significantly positively correlated.

## III. DATASETS

The MNLI dataset has been machine-translated into 14 different languages to produce synthetic training data in these languages. It is done to compute and analyze various benchmarks such as TRANSLATE-TRAIN and TRANSLATE-TEST (Conneau et al., 2020). While going through the dataset translation in the Hindi language, there has been a variety of discrepancy found. A few examples have been recorded in Fig. 1. We can see in example 1 where the original Hindi translation portrays a different meaning than what is being conveyed by the English counterpart. In contrast, if we look at the new Hindi translation, the sentence is appropriately translated. Example 2 and example 4 have no fundamental structure and meaning in the original Hindi translation, and the new Hindi translation gives the real meaning. In examples 5, 6 and 7, the premise/hypothesis in German, French and Swahili is recorded along with their corresponding original and new translations. We could find significant differences in each of the new translations compared to the original ones. Similarly, throughout 3,92,000 examples in the training set, there are excessive occurrences of mistranslations, which in turn lead to inaccurate benchmarks.

As we look deeper into every machine-translated train set, we find that the current cloud translation systems, such as Google Translate, perform at a level far better than that of three years ago. In releases since 2019, the average BLEU scores of Google Translate API over 100+ languages have improved by 5 pts and 7 pts on low-resource languages. We translated the English train set into all 14 languages using Google Translate and curated a separate dataset for each language. We also explored the quality of translation in the XNLI test and dev sets. Human translators have done these translations, but we still found considerable mistranslations. The new translations performed on Google Translate provide a much more accurate interpretation of their English counterparts, and therefore, a completely new dataset has been curated for each of the 15 languages in XNLI for test and dev sets. We call this new dataset XNLI 2.0, which consists of machine-translated train sets and test and dev sets on which evaluations will be performed.

## IV. EXPERIMENTATION AND RESULTS

### A. Training details

As already stated, this paper aims to do a comparative study of how the changes in dataset translations bring about changes in several benchmarks. We aim to show that (i) languages other than English could also be a better choice for natural language inference (NLI) and (ii) try to find languages that give better performance in understanding low-resource languages such as Swahili and Urdu. We primarily use XLM-RoBERTa (base), a model pre-trained on 2.5TB of filtered CommonCrawl data consisting of 100 languages (Conneau et al. 2020). Because of limited resources, it was not possible to reproduce the model from the original paper (Conneau et al., 2020), and therefore we used our training parameters.

For cross-lingual understanding, we first fine-tune XLM-RoBERTa(base) model on the MNLI dataset for three epochs. We use the Adam optimizer and the XLM-R tokenizer to encode the input dataset. 80% of the dataset is used for training while the other 20% is used for validation. After the model is trained, we first test it on the original test set of 14 languages and tabulate the accuracy score in Table I. Then, we test the model on the new test dataset of XNLI 2.0 and study how the changes in the test dataset produce changes in accuracy. The recorded accuracy on the original XNLI test dataset is approximately 73% for the model trained in English. In contrast, on our XNLI 2.0 test set, we obtain around a 3% increase in accuracy, i.e., ~76%.

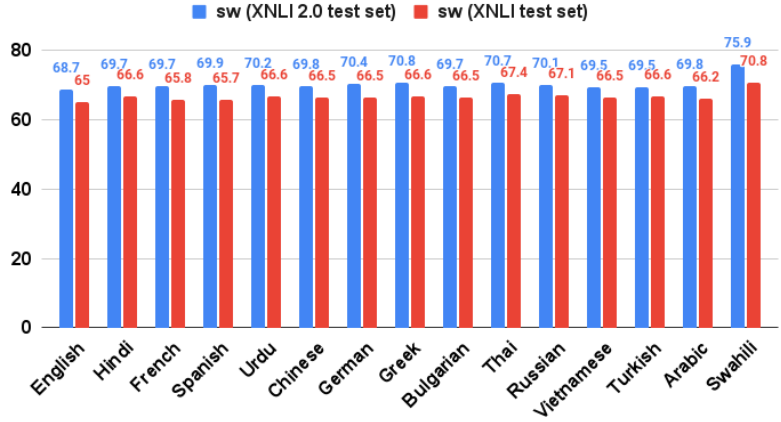

Fig. 3. Performance of models trained in 15 different languages on Swahili test set of XNLI 2.0.

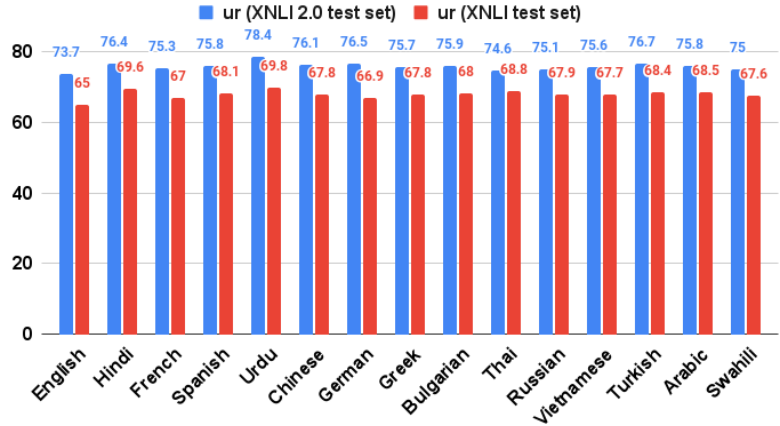

Fig. 4. Performance of models trained in 15 different languages on Urdu test set of XNLI 2.0.

Similarly, we trained separate models on machine-translated datasets of all other 14 languages and tested it on both XNLI and XNLI 2.0 datasets. We tabulate the accuracy score and present it in Table I and Table II.

### B. Analysis and Results

We have observed in all works related to XNLI that there is a heavy dependence on the English language dataset for training the model and performing cross-lingual transfer learning. This, in turn, limits our capability to not even test models trained in languages other than English. While English is a high-resource language, so are German, French, and Spanish.

Comparing the performance of our model on XNLI and XNLI 2.0 in Table I and Table II, we observe a difference in average accuracy in the range of 2.5%- 3% for all 15 languages. Because of fresh translations of the training dataset, there has been a significant improvement in the accuracy of the model. If we also analyze the model's performance in every language on the new test set, we can see that the model trained in German has an average accuracy of 78.15%. In comparison, the model trained in English has an average accuracy of 76.63%, which depicts that the model trained in German outperforms that trained in English in the cross-lingual natural language inference task.

Fig. 2 shows the performance of 15 different models (trained on each of the 15 languages) on every language in the XNLI 2.0 test set. To perform several tasks on models trained in these languages, we need to look at the performance of models on every language in test set and find the one that gives a better accuracy. This will facilitate in choosing a high-resource language in which a model could be trained and that gives a better performance in XLU. We can see in Fig. 2 that there are several high resource languages that give competent cross-lingual performance as compared to the same language in which the model is trained, such as German, Bulgarian and Spanish.

Table II gives an interesting viewpoint on languages that perform the best with regard to low-resource languages. In the case of Swahili, three languages perform the best and have almost similar accuracy. Thai (70.7%), Greek (70.8%), and German (70.4%) outperform English (68.7%), as can be seen in Fig. 3. In the case of Urdu, Turkish (76.7%) and German (76.5%) performs the best and outperform English (73.7%), as demonstrated in Fig. 4. One commonality worth noticing is how German (high-resource) language could be leveraged for performing tasks in low-resource languages instead of relying on English-only models.

## Conclusion

This paper demonstrates that our new machine-translated XNLI corpus (XNLI 2.0) performs around 3% better on cross-lingual natural language inference than the original machine-translated dataset (XNLI).

Although it was not possible for us to replicate the original model (Conneau et al., 2020), we still outperformed the original model using our XNLI 2.0 dataset, which establishes the fact that our XNLI 2.0 dataset should significantly exceed the original dataset performance if used in the original model.

We have also shown that there are many languages other than English, such as German, Greek, and Bulgarian on which models could be trained that perform well on cross-lingual language understanding tasks. Finally, we have shown how the high-resource languages perform better cross-lingual understanding of low-resource languages on the XNLI 2.0 dataset as compared to the original XNLI dataset which demonstrates that the re-translation of original XNLI dataset to XNLI 2.0 dataset has significantly improved the performance.

Our work in this paper establishes the possibility of cross-lingual transfer learning playing a significant role in several other tasks such as question-answering, text-summarization, and named-entity recognition, in which training language doesn't have to be restricted only to English, but several other high-resource languages could be put into use. Performing several natural language understanding tasks, especially in low-resource languages, will foster technological inclusion. Future works could include expanding the scope of the training dataset by including examples in several other low-resource languages such as Bengali, Kannada, Tamil, etc.

## Code and Data

The new dataset, i.e., XNLI 2.0, is available on the Hugging Face hub.

## Acknowledgements

We are very thankful to Dr. Naidila Sadashiv and Dr. Naveen N C for their invaluable guidance and mentoring throughout the research.